\documentclass[letterpaper, 10 pt, conference]{ieeeconf}
\IEEEoverridecommandlockouts
\usepackage{cite}
\usepackage{amsmath,amssymb,amsfonts}
\usepackage{algorithmic}
\usepackage{graphicx}
\usepackage{textcomp}
\usepackage{xcolor}
\usepackage{todonotes}
\usepackage{hyperref}
\hypersetup{
    colorlinks=true,
    linkcolor=blue,
    filecolor=magenta,      
    urlcolor=magenta,
    pdftitle={Overleaf Example},
    pdfpagemode=FullScreen,
    }

\usepackage[protrusion=true,
             expansion=true,
             tracking=true,
             final,
             stretch=25,factor=1250]{microtype}

\def\BibTeX{{\rm B\kern-.05em{\sc i\kern-.025em b}\kern-.08em
    T\kern-.1667em\lower.7ex\hbox{E}\kern-.125emX}}
\begin{document}

\title{%To BEV or not to BEV: 
 \LARGE \bf Translating Images into Maps
}

% \author{\IEEEauthorblockN{Avishkar Saha$^{1}$, Oscar Mendez$^{1}$, Chris Russell$^{2}$, Richard Bowden$^{1}$}
\author{{Avishkar Saha$^{1}$, Oscar Mendez$^{1}$, Chris Russell$^{2}$, Richard Bowden$^{1}$}
\thanks{$^{1}$Centre for Vision Speech and Signal Processing, University of Surrey, Guildford, UK, \{a.saha, o.mendez, r.bowden\}@surrey.ac.uk}
\thanks{$^{2}$\textit{Amazon}, Tubingen, Germany. cmruss@amazon.com}
}
\maketitle

\begin{abstract}
We approach instantaneous mapping, converting images to a top-down view of the world, as a translation problem. We show how a novel form of transformer network can be used to map from images and video directly to an overhead map or bird's-eye-view (BEV) of the world, in a single end-to-end network. We assume a 1-1 correspondence between a vertical scanline in the image, and rays passing through the camera location in an overhead map. This lets us formulate map generation from an image as a set of sequence-to-sequence translations.  Posing the problem as translation allows the network to use the context of the image when interpreting the role of each pixel. This constrained formulation, based upon a strong physical grounding of the problem, leads to a restricted transformer network that is convolutional in the horizontal direction only. The structure allows us to make efficient use of data when training, and obtains state-of-the-art results for instantaneous mapping of three large-scale datasets, including a 15\% and 30\% relative gain against existing best performing methods on the nuScenes and Argoverse datasets, respectively. %We make our code available on \url{https://github.com/avishkarsaha/translating-images-into-maps}.
\end{abstract}
\vspace{-0.2cm}
% \begin{IEEEkeywords}
% 3D scene understanding, bird-eye-view estimation, semantic segmentation
% \end{IEEEkeywords}

\section{Introduction}
Many tasks in autonomous driving 
%and planning
are substantially easier from a top-down, map or bird's-eye view (BEV). As many autonomous agents are restricted to the ground-plane, an overhead map is a convenient low-dimensional representation, ideal for navigation, that captures relevant obstacles and hazards.  
For scenarios such as autonomous driving, semantically segmented BEV maps must be generated on the fly as an instantaneous estimate, to cope with freely moving objects and scenes that are visited only once.

Inferring 
%semantic 
BEV maps from images requires determining the correspondence between image elements and their location in the world. Multiple works guide their transformation with dense depth and image segmentation maps 
%either as part of their model's input 
\cite{sengupta2012automatic, pan2020cross, liu2020understanding, wang2019parametric, schulter2018learning}, 
%or as part of its training signal \cite{schulter2018learning}.Recently, several works
while others \cite{lu2019monocular, mani2020monolayout, Roddick_2020_CVPR, philion2020lift, saha2021enabling} have developed approaches which resolve depth and semantics implicitly. 
%learning only from the image and its BEV ground truth.
Although some exploit the camera's geometric priors \cite{Roddick_2020_CVPR, philion2020lift, saha2021enabling}, they do not explicitly learn the interaction between image elements and the BEV-plane. 

Unlike previous approaches, we treat the transformation to BEV as an image-to-world translation problem, where the objective is to learn an alignment between vertical scan lines in the image and polar rays in BEV. The projective geometry therefore becomes implicit to the network. For our alignment model, we adopt transformers \cite{vaswani2017attention}, an attention-based architecture for sequence prediction. With its attention mechanisms, we explicitly model pairwise interactions between vertical scanlines in the image and their polar BEV projections.  Transformers are well-suited to the image-to-BEV transformation problem, as they can reason about interdependence between objects,  depths and the lighting of the scene to achieve a globally consistent representation.

We embed our transformer-based alignment model within an end-to-end learning formulation which takes as input a monocular image with its intrinsic matrix, and predicts semantic BEV maps for static and dynamic classes. 

The contributions of our paper are (1) We formulate  generating a  BEV map from an image as a set of 1D sequence-to-sequence translations. (2) By physically grounding
our formulation we construct a restricted data-efficient transformer network that is convolutional with respect to the horizontal x-axis, yet spatially-aware. (3) By combining our formulation with 
%recent progress on 
\emph{monotonic attention} from the language domain, we show that knowledge of what is below a point in an image is more important than knowledge of what is above it for accurate mapping; although using both leads to best performance. (4) We show how axial attention improves performance by providing temporal awareness and demonstrate state-of-the-art results across three large-scale datasets.

% We provide an overview of prior work relating to BEV estimation from images and encoder-decoder transformer architectures in Sec. \ref{sec:related_work}. Our methodology in Sec. \ref{sec:method} formulates image-to-BEV as a translation problem along with our model architecture. Finally in Sec. \ref{sec:experiments} we present our results before concluding in Sec \ref{sec:conclusion}.

%------------------------------------- RELATED WORK -----------------------------------------%
\vspace{-0.1cm}
\section{Related Work} \label{sec:related_work}
\textbf{BEV object detection:} 
%Initial work by Palazzi \textit{et al.} \cite{palazzi2017learning}  generated a BEV representation of dynamic objects in a scene, with 2D object detection in the image-plane followed by mapping to BEV. 
Early approaches detected objects in the image and then regressed 3D pose parameters \cite{wang2019monocular, mousavian20173d, kehl2017ssd, simonelli2019disentangling, poirson2016fast, palazzi2017learning}. The Mono3D \cite{chen2016monocular} model instead  generated 3D bounding box proposals on the ground plane and scored each one by projecting into the image. However, all these works lacked global scene reasoning in 3D as each 
%3D bounding box 
proposal was generated independently. OFTNet \cite{roddick2018orthographic} overcame this by generating 3D features 
%derived 
from projecting a 3D voxel grid into the image, and performing 3D object detection over those features. While it reasons directly in BEV, the context available to each voxel depends upon its distance from the camera, in contrast, we decouple this relationship to allow each BEV position access to the entire vertical axis of the image. 

\textbf{Inferring semantic BEV maps:} BEV object detection has been extended to building semantic maps from images for both static and dynamic objects. Early work in road layout estimation  \cite{sengupta2012automatic} performed semantic segmentation in the image-plane and assumed a flat world mapping to the ground plane via a homography. However, as the flat world assumption leads to artifacts for dynamic objects such as cars and pedestrians,  
%Similarly, in road layout estimation from images, 
others \cite{wang2019parametric, liu2020understanding, schulter2018learning} exploit
%pixel-level 
depth and semantic segmentation maps to lift 
%segmented 
objects into BEV. While such intermediate representations provide strong priors, they require image depth and segmentation maps as  additional input. 

Several works instead reason about semantics and depth implicitly. Some use camera geometry to transform the image into BEV \cite{Roddick_2020_CVPR, philion2020lift, saha2021enabling} while others learn this transformation implicitly \cite{pan2020cross, lu2019monocular, mani2020monolayout}. Current state-of-the-art approaches can be categorised as taking a `compression' \cite{Roddick_2020_CVPR, saha2021enabling} or `lift' approach \cite{philion2020lift, fiery2021} to the transformation. `Compression' approaches vertically condense image features into a bottleneck representation and then expand out into BEV, thus creating an implicit relationship between an object's depth and the context available to it. This
%, in turn, 
increases its susceptibility to ignore small, distant objects.  `Lift' approaches instead expand each image into a frustum of features to learn a depth distribution for each pixel. However, each pixel is given the entire image as context, potentially increasing 
%its tendency for 
overfitting due to redundancies in the image. Furthermore, neither 
%compression nor lift 
approaches have spatial awareness, meaning they are unable to leverage the structured 
%spatial 
environments of urban scenes. We overcome issues with both these approaches by (1) maintaining the spatial structure of the image to explicitly model its alignment with the BEV-plane and (2) adding spatial awareness which allows the network to assign image context across the ray space based on both content and position.

\textbf{Encoder-decoder transformers:} Attention mechanisms were first proposed by Bahdanau \textit{et al.} \cite{bahdanau2015neural} for machine translation to learn an alignment between source and target sequences using recurrent neural networks (RNNs). Transformers, introduced by Vaswani \textit{et al.} \cite{vaswani2017attention}, instead implemented attention within an entirely feed-forward network, leading to
%Through self-attention, they eliminated the need for recurrence by allowing global computation across sequences.
%Their superiority over RNNs led to 
state-of-the-art performance in many tasks \cite{devlin2019bert, radfordlanguage}.

%Our use of transformers for parallel decoding differs from their original use as auto-regressive models \cite{sutskever2014sequence} in sequence transduction tasks \cite{vaswani2017attention}. As an autoregressive model, training requires causal masking in the decoder, where every predicted token in the output sequence can only be conditioned on previous tokens in the sequence. At test time, this results in an auto-regressive decoding where each prediction is passed back into the decoder. In contrast, our output is a prediction along a spatial axis and we reason globally across it. For this, we exploit the feed-forward, parallel nature of transformers: we omit causal masks during training, and keep the model entirely feed-forward at test time with parallel decoding. 

Like us, the 2D detector DETR \cite{carion2020detr} performs decoding in a spatial domain through attention. 
However, their predicted output sequences are sets of object detections, which have no intrinsic order to them, and permits the use of attention's permutation invariant nature without any spatial awareness. In contrast, the order of our predicted BEV ray sequences  is inherently spatial and so we need spatial awareness and therefore permutation equivariance in our decoding.

%-----------------------------------------MODEL-----------------------------------------------%
\begin{figure*}[t]
\centerline{\includegraphics[clip,trim=0 225 0 0,width=0.85\linewidth]{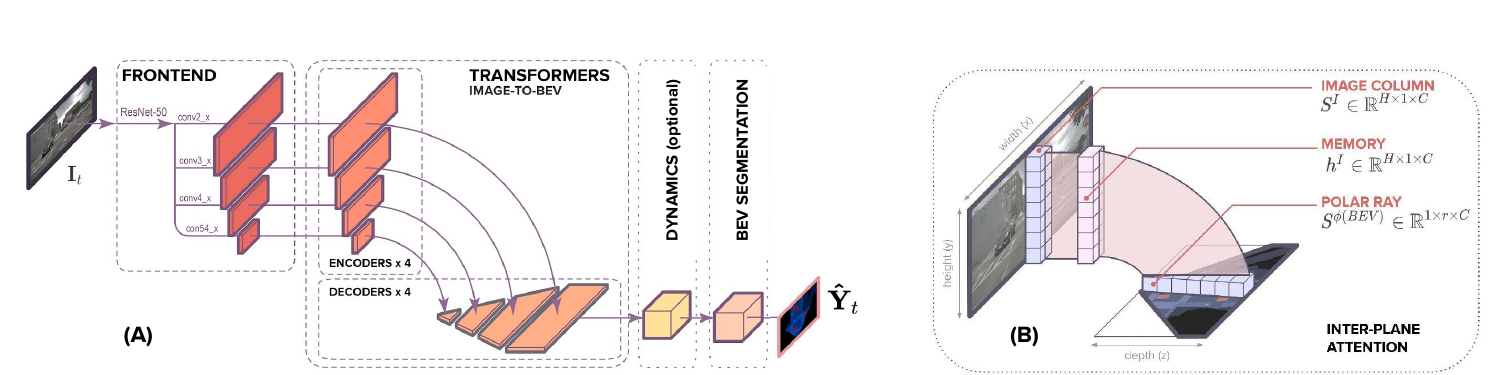}}
\vspace{-0.2cm}
\caption{(A) Our model architecture. The \textbf{Frontend} extracts spatial features at multiple scales. \textbf{Encoder-decoder transformers} translate  spatial features from the image to BEV. An optional \textbf{Dynamics Module} uses past spatial BEV features to learn a spatiotemporal BEV representation. A \textbf{BEV Segmentation Network} processes the BEV representation to produce multi-scale occupancy grids. (B) Our inter-plane attention mechanism. In our attention-based model, vertical scan lines in the image are passed one by one to a transformer encoder to create a `memory' representation which is  decoded into a BEV polar ray.}
\label{fig:model_arch}
\vspace{-0.5cm}
\end{figure*}

\section{Method} \label{sec:method}
Our goal is to learn a model $\mathbf{\Phi}$ that takes a monocular image $\mathbf{I}$ and produces a semantically segmented birds-eye-view map of the scene $\mathbf{Y}$. Formally, given an input image $\mathbf{I} \in \mathbb{R}^{3 \times H \times W}$ and its intrinsic matrix $\mathbf{C} \in \mathbb{R}^{3 \times 3}$, our model predicts a set of binary variables $\mathbf{Y}^k \in \mathbb{R}^{X \times Z}$ for each class $k \in K$:
\begin{equation}
    p(\mathbf{Y}^k|\mathbf{I}, \mathbf{C}) = \mathbf{\Phi}(\mathbf{I}, \mathbf{C}),
\end{equation}
where $\Phi$ is a neural network 
%with weights $\theta$
trained to resolve both semantic and positional uncertainties. 

The design of our network rests on our novel transformation between the image-plane $\mathbb{P}^I$ and BEV-plane $\mathbb{P}^{BEV}$. 
%To learn this mapping, we use the fact that the ray a pixel lies along on the ground plane can be determined by camera geometry, and thus positioning an object in BEV can be reduced to predicting its depth. 
Our end-to-end approach, as shown in Fig.~\ref{fig:model_arch}a, is composed of the following subtasks: (1) constructing representations in the image-plane which encode semantics and some knowledge of depth, (2) transforming the image-plane representation to BEV and (3) semantically segmenting the BEV-representation. 
%\color{red} The novelty of our work is in the image-to-BEV transformation (2), with (1) and (3) being standard architectures used in vision. \color{black}
  
\subsection{Image-to-BEV Translation}
Transforming  from image to BEV requires a mapping  which determines the image pixel correspondence to BEV polar ray. As camera geometry dictates a 1-1 correspondence between each vertical scanline and its associated ray, we treat the mapping
%process 
as a set of sequence-to-sequence translations. With reference to Fig.~\ref{fig:model_arch}b, we want to find the discretized radial depths of elements in the vertical scan line of an image, up to $r$ metres from the camera: we have an image column $S^I \in \mathbb{R}^H$, and we want to find its BEV ray $S^{\phi(BEV)} \in \mathbb{R}^r$, where $H$ is the height of the column and $r$ represents the radial distance from the camera. This mapping can be viewed as an assignment of semantic objects from the image-plane to their positional slots along a ray in the BEV-plane. 

We propose learning the alignment between input scanlines and output polar rays through an attention mechanism \cite{bahdanau2015neural}. We employ attention in two ways: (1) \textit{inter-plane attention} as shown in Fig.\ref{fig:model_arch}b, which initially assigns features from a scanline to a ray and (2) \textit{polar ray self-attention} that globally reasons about its positional assignments across the ray. We motivate both uses below, starting with inter-plane attention.

%-----------------attention v compression diagram ----------------------------%
%\begin{figure*}[t]
%\centerline{\includegraphics[trim=0 140 0 180,clip,width=0.85\linewidth]{cvprdiagram2.pdf}}
%\caption{Image-to-BEV transformation-based approach: (left) Compression (right) Our inter-plane attention mechanism. In our attention-based model, vertical scan lines in the image are passed one by one to a transformer encoder to create a `memory' representation which is  decoded into a BEV polar ray.}
%\label{fig:model_arch}
%\vspace{-0.5cm}
%\end{figure*}

% start from an ideal world, but we dont have perfect segmentation or depth, and the amount of context required varies along the ray
\textbf{Inter-plane attention:} Consider a semantically segmented image column and its corresponding polar BEV ground truth. Here, alignment between the column and the ground truth ray is `hard', \emph{i.e.} each pixel in the polar ray corresponds to a single semantic category from the image column. Thus, the only uncertainty that must be resolved to make this a hard-assignment is the depth of each pixel. However, when making this assignment, we need to assign features that aid in resolving semantics and depth. Hence, a hard assignment would be detrimental. Instead, we want a soft-alignment, where every pixel in the polar ray is assigned a combination of elements in the image column, \emph{i.e}. a \textit{context} vector. Concretely, when generating each radial element $S^{\phi(BEV)}_i$, we want to give it a \textit{context} $c_i$ based on a convex combination of elements in the image column $S^{I}$ and the radial position $r_i$ of the element $S^{\phi(BEV)}_i$ along the polar ray. This need for context assignment motivates our use of soft-attention between the image column and its polar ray, as illustrated in Fig.~\ref{fig:model_arch}.

Formally, let $\mathbf{h} \in \mathbb{R}^{H \times C}$ represent the encoded ``memory'' of an image column of height $H$, and let $\mathbf{y} \in \mathbb{R}^{r \times C}$ represent a \textit{positional query} which encodes relative position along a polar ray of length $r$. %Our aim is to produce
We  generate a context $\mathbf{c}$ based on the input sequence $\mathbf{h}$ and the query $\mathbf{y}$ through alignment $\boldsymbol{\alpha}$ between elements in the input sequence and their radial position. First, the input sequence $\mathbf{h}$ and positional query $\mathbf{y}$ are projected by matrices $W_Q \in \mathbb{R}^{C \times D}$ and $W_K \in \mathbb{R}^{C \times D}$ to the corresponding representations $Q$ and $K$:
\begin{equation}\label{eq:project}
\begin{split}
    Q(\mathbf{y}_i) = \mathbf{y}_i W_Q   \\
    K(\mathbf{h}_i) = \mathbf{h}_i W_K . \\
\end{split}
\end{equation}
Following common terminology, we refer to $Q$ and $K$ as `queries' and `keys' respectively. After projection, an unnormalized alignment score $e_{i,j}$ is produced between each memory-query combination using the scaled-dot product \cite{vaswani2017attention}:
\begin{equation}\label{eq:energy}
    e_{i,j} = \frac{\langle Q(\mathbf{y}_i), K(\mathbf{h}_j) \rangle}{\sqrt{D}}.
\end{equation}
The energy scalars are then normalized using a $\mathrm{softmax}$ to produce a probability distribution over the memory: 
\begin{equation}\label{eq:alpha}
    \alpha_{i,j} = \frac{\mathrm{exp}(e_{i,j})}{\sum_{k=1}^{H}\mathrm{exp}(e_{i,k})}.
\end{equation}
Finally, the context vector is computed as a weighted sum of $K$:
\begin{equation}\label{eq:context}
    c_i = \sum_{j=1}^H \alpha_{i,j}K(\mathbf{h}_j).
\end{equation}
Generating the context this way allows each radial slot $r_i$ to independently gather relevant information from the image column; and represents an initial assignment of components from the image to their BEV locations. Such an initial assignment is analogous to lifting a pixel based on its depth. However, it is lifted to a distribution of depths and thus should be able to overcome common pitfalls of sparsity and elongated object frustums. This means that the image-context available to each radial slot is decoupled from its distance to the camera. Finally, to generate BEV feature $S^{\phi(BEV)}_i$ at radial position $r_i$, we globally operate on the assigned contexts for \emph{all} radial positions $\mathbf{c}=\{c_1, ..., c_r\}$:
\begin{equation}
    S^{\phi(BEV)}_i = g(\mathbf{c}), 
\end{equation}
where $g(.)$ is a nonlinear function reasoning across the \emph{entire} polar ray. We describe its role below. 

\textbf{Polar ray self-attention:} The need for the non-linear function $g(.)$ as a global operator arises out of the limitations brought about by generating each context vector $c_i$ independently. Given the absence of global reasoning for each context $c_i$, the spatial distribution of features across the ray is unlikely to be congruent with object shape, locally or globally. Rather, this distribution may only represent scattered suggestions of object-part positions. Therefore, we need to operate globally across the ray to allow the assigned scanline features to reason about their placement within the context of the entire ray, and thus aggregate information in a manner that generates coherent object shapes.

Global computation across the polar ray is computed much like soft-attention outlined in Eq.~\eqref{eq:project} - ~\eqref{eq:context}, except that the self-attention is applied to the ray only.  Eq.~\eqref{eq:project} is recalculated with a new set of weight matrices with inputs to both equations replaced with the context vector $c_i$. 

\textbf{Extension to transformers:}
Our inter-plane attention can be extended to attention between the encoder-decoder of transformers by replacing the key $K(\mathbf{h}_j)$ in Eq.~\eqref{eq:context} with another projection of the memory $\mathbf{h}$, the `value'. Similarly, polar-ray self-attention can be placed within a transformer-decoder by replacing the key in Eq.~\eqref{eq:context} with a projection of the context $c_i$ to represent the value.

%-------------Monotonic Attention------------------------------%
\subsection{Infinite lookback monotonic attention}\label{sec:mono_attention}
Although soft-attention is sufficient for learning an alignment between any arbitrary pair of source-target sequences, our sequences exist in the physical world where the alignment exhibits physical properties based on their spatial ordering. Typically, in urban environments, depth monotonically increases with height \emph{i.e.,}\ as you move up the image, you move further away from the camera. We enforce this through monotonic attention with infinite lookback \cite{arivazhagan2019monotonic}. This constrains radial depth intervals to observe elements of the image column that are monotonically increasing in height but also allows context from the bottom of the column (or equivalently, previous memory entries). 

Monotonic attention (MA) was originally proposed for computing alignments for simultaneous machine translation \cite{raffel2017online}. However, the `hard' assignment between source and target sequence means important context is neglected. This led to the development of MA with infinite lookback (MAIL) \cite{chiu2018monotonic, arivazhagan2019monotonic, ma2019monotonic}, which combined hard MA with soft-attention that extends from the hard assignment to the beginning of the source sequence. We adopt MAIL as a way of constraining our attention mechanism to potentially prevent overfitting by ignoring the redundant context in the vertical scan line of an image. The primary objective of our adoption of MAIL is to understand whether context below a point in an image is more helpful than what is above.

 We employ MAIL by first calculating a hard-alignment using monotonic attention. This makes a hard assignment of context $c_i$ to an element of the memory $\mathbf{h}_j$, after which a soft-attention mechanism over previous memory entries $\mathbf{h}_1, ..., \mathbf{h}_{j-1}$ is applied. \color{black} Formally, for each radial position $\mathbf{y}_i \in \mathbf{y}$ along the polar ray, the decoder begins scanning memory entries from index $j=t_{i-1}$, where $t_i$ is the index of the memory entry chosen for position $\mathbf{y}_i$. For each memory entry, it produces a selection probability $p_{i,j}$, which corresponds to the probability of either stopping and setting $t_i=j$ and $c_i = \mathbf{h}_{t_i}$, or moving onto the next memory entry $j+1$. As  hard assignment is not differentiable, training is instead carried out with respect to the expected value of $c_i$, with the monotonic alignments $\alpha_{i,j}$ calculated as follows:
\begin{equation} \label{eq:ma_prob}
    p_{i,j} = \mathrm{sigmoid}(\mathrm{Energy}(\mathbf{y}_i, \mathbf{h}_j)),
\end{equation}
\begin{equation}\label{eq:ma_alpha}
    \alpha_{i,j} = p_{i,j} \left((1-p_{i,j-1})\frac{\alpha_{i,j-1}}{p_{i,j-1}} + \alpha_{i-1,j} \right),
\end{equation}
where the $\mathrm{Energy}$ function is calculated in the same manner as Eq.~\eqref{eq:energy}.
Assuming monotonic attention stops at $t_i$, the infinite lookback strategy first computes energies $e_{i,k}$ using equation Eq. \ref{eq:energy} for $k \in 1, 2, ..., t_i$. The attention distribution over the allowed states is  calculated as follows:
\begin{equation}
    \beta_{i,j} = \sum_{k=j}^{H} \left(\frac{\alpha_{i,k}\mathrm{exp}(e_{i,k})}{\sum_{l=1}^k\mathrm{exp}(e_{i,l})}\right).
\end{equation}
This effectively represents a distribution over image-elements which lie below a point in the image; to calculate a distribution over only what lies above a point in the image, the image column can be flipped.
The context vector is calculated similar to inter-plane attention, where $c_i = \sum_{j=1}^H \beta_{i,j}K(\mathbf{h}_j)$.

%--------- Model Architecture ---------%

\subsection{Model architecture}\label{sec:model_arch}
We build an architecture that facilitates our goal of predicting a semantic BEV map from a monocular image around this alignment model. As shown in Fig.~\ref{fig:model_arch}, it contains three main components: a  standard CNN backbone which extracts spatial features in the image-plane, encoder-decoder transformers to translate features from the image-plane to BEV and finally a segmentation network which decodes BEV features into semantic maps. 

\textbf{2D Multi-scale feature learning in $\mathbb{P}^I$:} Reconstructing an image in BEV requires representations which can detect scene elements at varying depths and scale. Like prior object detection methods \cite{tian2019fcos, Roddick_2020_CVPR, saha2021enabling}, we handle this scale variance using a CNN backbone with a feature pyramid to produce feature maps $\mathbf{f}_{t,s}^I \in \mathbb{R}^{C \times h_s \times w_s}$ at multiple scales $u \in U$.

\textbf{1D Transformer encoders in $\mathbb{P}^I$:}
This component encodes long-range vertical dependencies across the input features through self-attention, using an encoder for each scale $u$ of features (second left block of Fig.~\ref{fig:model_arch}a). Each scale of features $\mathbf{f}_{t, u}^I$ is first reshaped into its individual columns, creating $w_u$ sequences of length $h_u$ and dimension $C$. Each encoder layer has a standard architecture consisting of multi-head attention and a feed forward network. Given the permutation invariance of the transformer, we add fixed 1D sinusoid positional encodings \cite{vaswani2017attention} to the input of each attention layer. The $U$ encoders each produce a memory $\mathbf{h}_{t,u}^I \in \mathbb{R}^{w_u \times h_u \times C}$.

\textbf{1D Transformer decoders in $\mathbb{P}^{BEV}$:} This component generates independent sequences of BEV features along a polar ray through multi-head attention across the encoder memory.  As shown in the second left block of Fig.~\ref{fig:model_arch}, there is one transformer decoder for each transformer encoder. Every encoded image column $\mathbf{h}^I \in \mathbb{R}^{h_u \times C}$ is transformed to a BEV polar ray $\mathbf{f}^{\phi(BEV)} \in \mathbb{R}^{r_u \times C}$, where $r_u$ is the radial distance along the ray. Given the desired output sequence of length $r_u$, the decoder takes in $r_u$ positional embeddings, which we refer to as \textit{positional queries}. These are $r_u$ unique embeddings with fixed sinusoid positional information added to them, just like our encoder above. When replacing the encoder-decoder multi-head soft-attention with monotonic attention, each head in the decoder is replaced with a monotonic attention head from Eq.~\eqref{eq:ma_alpha}. The $U$ decoders each output $w_u$ BEV sequences of length $r_u$ along the polar ray, producing a polar encoding $\mathbf{f}^{\phi(BEV)} \in \mathbb{R}^{w_u \times r_u \times C}$. Similar to prior work which builds stixel representations from an image \cite{badino2009stixel, pfeiffer2011modeling}, each image column in our model corresponds to an angular coordinate in the polar map. Finally we concatenate along the ray to obtain a single 2D polar feature map and convert to a rectilinear grid, to create our BEV representation $\mathbf{f}^{{BEV}}_{t} \in \mathbb{R}^{C \times Z \times X}$. 

Our transformer encoder and decoder use the same set of projection matrices for every sequence-to-sequence translation, giving it a structure that is convolutional along the $x$-axis and allowing us to make efficient use of data when training. We constrain our translations to 1D sequences as opposed to using the entire image to make learning easier, a decision we analyze in section \ref{sec:ablations}.

\textbf{Polar-adaptive context assignment:} 
The positional encodings applied to the transformer 
%encoder and decoder 
so far have all been 1D. While this allows our convolutional transformer to leverage spatial relationships between height in the image and depth, it remains agnostic to polar angle. However, the angular domain plays an important role in urban environments. For instance, images display a broadly structured distribution of object classes across their width (e.g. pedestrians are typically only seen on sidewalks, which lie towards the edges of the image). Furthermore, object appearance is also structured along the width of the image as they are typically orientated along orthogonal axes and viewing angle changes its appearance. To account for such variations in 
% object 
appearance and distribution across the 
%width of the 
image, we add additional positional information by encoding polar angle in our 1D scanline-to-ray translations.

%While positional information is needed to make the attention mechanisms of the transformer permutation equivariant, it plays another particularly important role for our task as we operate in highly structured urban environments. 

%The relationship between height and depth can be learnt through the addition of 1D positional information to our scanline and ray sequences. This imposes inductive biases which aids the distribution of scanline elements along the ray, making our convolutional transformer spatially aware along the height and depth axis, but agnostic to polar-angle.

%However, there is another relationship that exists in our scenario. Given the viewpoint of the camera is always fixed in self-driving vehicles, the distribution of object pose relative to the camera varies according to its polar angle. Again, this structure can be exploited to create a polar-adaptive transformer, where additional positional information is added to each sequence and ray based on polar angle. Ultimately this would allow the network to use the same set of project matrices for each sequence to sequence translation, but modulate each one based on its spatial coordinates.
\color{black}

\textbf{Dynamics with axial attention in $\mathbb{P}^{BEV}$:}
This component incorporates temporal information from past estimates to build a spatiotemporal BEV representation of the present. As the representations built by the previous components are entirely spatial, we add a simple component based on axial attention to make the model temporally aware. The placement of this optional module can be seen in Fig.~\ref{fig:model_arch}a. 
%where it sits after the \emph{resample} block. 
We obtain BEV features for multiple timesteps, creating a representation $\mathbf{f}_{1:t}^{BEV} \in \mathbb{R}^{T \times C \times Z \times X}$. We apply axial-attention across the spatial and temporal axes, giving every pixel at every timestep axial context from the other timesteps. Our temporal aggregation means the features of any timestep now contain dynamics across the sequence, and the module can use any of these features in its forward pass. This module is optional as it  builds a spatiotemporal representation.
%which incorporates temporal information across multiple timesteps
It can be omitted when constructing a purely spatial model.

\textbf{Segmentation in $\mathbb{P}^{BEV}$:} To decode our BEV features into semantic occupancy grids, we adopt a convolutional encoder-decoder structure used in prior segmentation networks \cite{yu2018deep, saha2021enabling}. The aggregated module structure (right block of Fig.~\ref{fig:model_arch}a), 
takes BEV features $\mathbf{f}^{{BEV}}_{t} \in \mathbb{R}^{C \times Z \times X}$ and outputs occupancy grids $\mathbf{m}^{BEV}_{t,u} \in \mathbb{R}^{classes \times x_u \times z_u}$ for scales $u \in U$. Moving from the 1D attention mechanisms of our transformer to the two-dimensional locality of convolutions provides contextual reasoning across the horizontal $x$-axis which helps stitch together potential discontinuities between adjacent polar rays and their subsequent rectilinear resampling. 

\textbf{Loss in $\mathbb{P}^{BEV}$:} As the training signal provided to the predicted occupancy grids must resolve both semantic and positional uncertainties, we use the same multi-scale Dice loss as \cite{saha2021enabling}. At each scale $u$, the mean Dice Loss across classes $K$ is:
\begin{equation}
    \mathcal{L}^u = 1 - \frac{1}{|K|} \sum_{k=1}^K \frac{2 \sum_i^N \hat{y}_i^k y_i^k}{\sum_i^N \hat{y}_i^k + y_i^k + \epsilon},
\end{equation}
where $y_i^k$ is the ground truth binary variable grid cell, $\hat{y}_i^k$ the predicted sigmoid output of the network, and $\epsilon$ is a constant used to prevent division by zero.

%\begin{table}
%\vspace{-0.2cm}
%\centering
%\caption{IoU(\%) 
%across object class groups 
%for monotonic attention mechanisms.}
%\label{tab:looking_dir}
%\begin{tabular}{cccc}
%\hline
%View direction & Static classes  & Dynamic classes & Mean \\ \hline
%Looking down             & 29.5	& 15.8	& 22.1                      \\
%Looking up             & 29.9 &	17.1 &	23.0                  \\
%Looking both ways               & \textbf{32.4}           & \textbf{19.4}   & \textbf{25.0}            \\ \hline
%\end{tabular}
%\vspace{-0.3cm}
%\end{table}

%-------------- SOTA IoU comparison -------------%

\begin{table*}[t]
\centering
\caption{IoU(\%) on the nuScenes validation split and baseline results of \cite{Roddick_2020_CVPR}.\vspace{-0.2cm}}
\label{tab:soa_comparison}
\resizebox{0.95\linewidth}{!}{%
\begin{tabular}{c|cccccccccccccc|c}
\hline
\scriptsize Method          &\scriptsize  Drivable            &\scriptsize  Crossing          &\scriptsize  Walkway             &\scriptsize  Carpark             &\scriptsize  Bus                 &\scriptsize  Bike             &\scriptsize  Car       &\scriptsize  Cons.Veh.          &\scriptsize  Motorbike         &\scriptsize  Trailer             &\scriptsize  Truck               &\scriptsize  Ped.        &\scriptsize   Cone         &\scriptsize  Barrier             &\scriptsize  Mean                \\ \hline
\scriptsize  VED \cite{lu2019monocular}            & 54.7                & 12.0                & 20.7                & 13.5                & 0.0                 & 0.0                 & 8.8       & 0.0          & 0.0                & 7.4                 & 0.2                 & 0.0                & 0                  & 4.0                 & 8.7                 \\
\scriptsize VPN \cite{pan2020cross}             & 58.0                & 27.3                & 29.4                & 12.3                & 20.0                & 4.4                 & 25.5        & 4.9        & 5.6                & {\textbf{16.6}} & 17.3                & 7.1                & 4.6                & 10.8                & 17.5                \\
\scriptsize  PON \cite{Roddick_2020_CVPR}             & 60.4                & 28.0                & 31.0       & 18.4                & 20.8                & 9.4                 & 24.7        & 12.3        & 7.0                & {\textbf{16.6}} & 16.3                & 8.2      & 5.7                & 8.1                 & 19.1                \\

\scriptsize STA-S \cite{saha2021enabling} & 71.1 & 31.5 & 32.0 & 28.0 & 22.8 & 14.6 & 34.6 & 10.0 & 7.1 & 11.4 & 18.1 & 7.4 & 5.8 & 10.8 & 21.8 \\

 \scriptsize Our Spatial & \textbf{72.6}	& \textbf{36.3} & \textbf{32.4} & \textbf{30.5} & \textbf{32.5} & \textbf{15.1} & \textbf{37.4} & \textbf{13.8} & \textbf{8.1} & 15.5 & \textbf{24.5} & \textbf{8.7} & \textbf{7.4} & \textbf{15.1} & \textbf{25.0} \\ \hline    
 
 \scriptsize STA-ST\cite{saha2021enabling} & 70.7 & 31.1 & 32.4 & \textbf{33.5} & 29.2 & 12.1 & 36.0 & 12.1 & \textbf{8.0} & 13.6 & 22.8 & 8.6 & 6.9 & 14.2 & 23.7 \\
 
\scriptsize Our Spatiotemp. & \textbf{74.5}	& \textbf{36.6}	& \textbf{35.9}	& 31.3	& \textbf{32.8}	& \textbf{14.7}	& \textbf{39.7}	& \textbf{14.2}	& 7.6	& \textbf{13.9}	& \textbf{26.3}	& \textbf{9.5}	& \textbf{7.6}	& \textbf{14.7} & \textbf{25.7} \\ \hline

\end{tabular}%
}
\vspace{-0.3cm}
\end{table*}
\color{black}
\begin{figure*}[t]
\centerline{\includegraphics[trim=0 330 0 0,clip,width=0.92\linewidth]{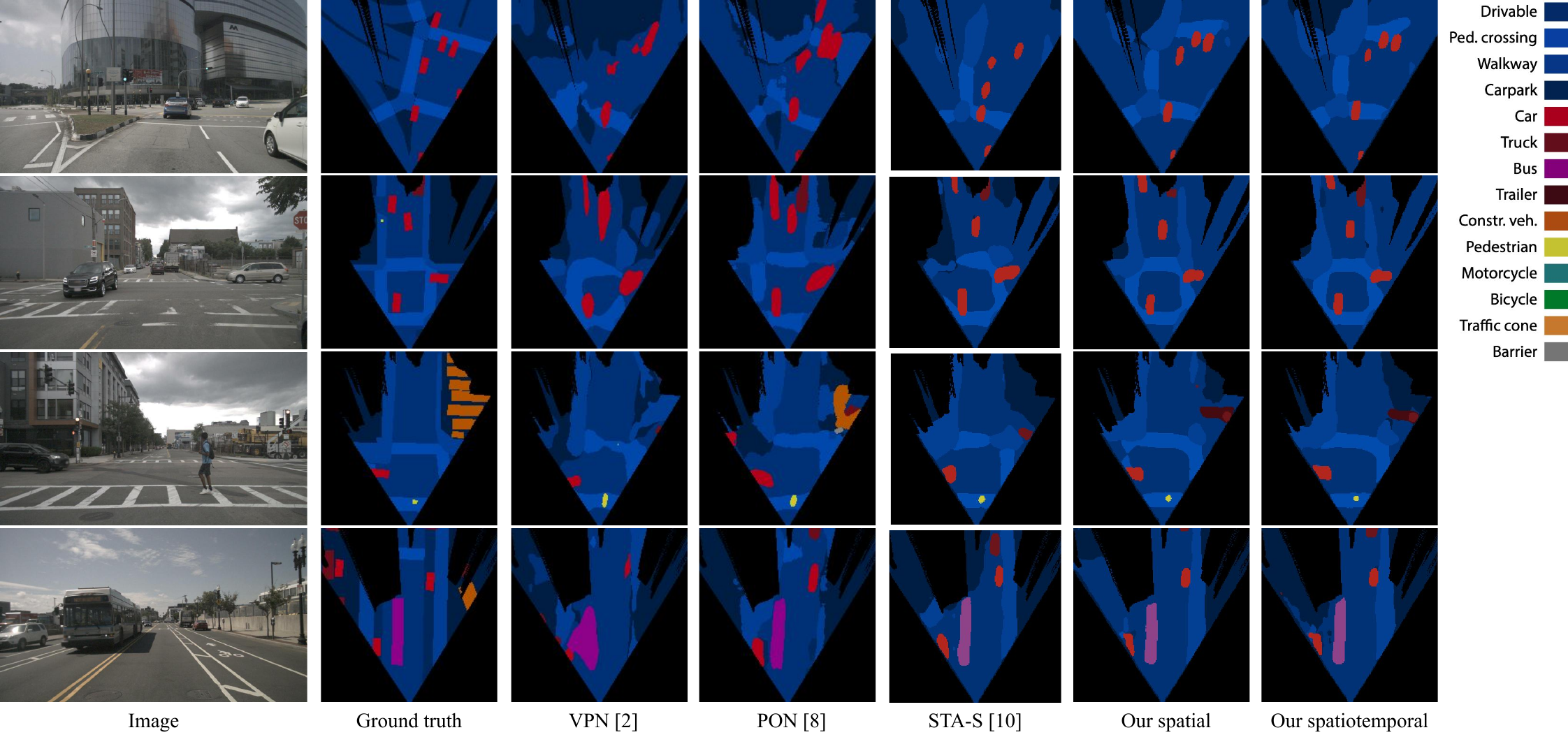}}
\vspace{-0.4cm}
\caption{Qualitative results on the nuScenes validation set of \cite{Roddick_2020_CVPR}. 
%Like the quantitative assessment, 
We compare against baseline results of prior work reported in \cite{Roddick_2020_CVPR} and follow their colour scheme. For fair comparison, we apply the ground truth visibility mask (black) to the predicted images as was done in \cite{Roddick_2020_CVPR}.}
\label{fig:soa_comp}
\vspace{-0.3cm}
\end{figure*}
%-----------------------------------------EXPERIMENTS-----------------------------------------------%
\section{Experiments and Results} \label{sec:experiments}
We evaluate the effectiveness of treating the image-to-BEV transformation as a translation problem on the nuScenes dataset \cite{caesar2020nuscenes}; with ablations on lookback direction in monotonic attention, the utility of long-range horizontal context and the effect of polar positional information. Finally, we compare our approach to current state-of-the-art approaches on the {\bf nuScenes \cite{caesar2020nuscenes}, Argoverse \cite{chang2019argoverse}} and {\bf Lyft \cite{kesten2019lyft}} datasets.

\textbf{Dataset:}
The nuScenes dataset \cite{caesar2020nuscenes} consists of 1000 20-second clips captured across Boston and Singapore, annotated with 3D bounding boxes and vectorized road maps.
We follow \cite{Roddick_2020_CVPR}'s data generation process, object classes and training/validation splits to allow fair comparison. We use nuScenes for our ablation studies as it is considerably larger and contains more object categories.

\textbf{Implementation:}
Our frontend uses a pretrained ResNet-50 \cite{he2016deep} with a feature pyramid \cite{lin2017feature} on top. BEV feature maps built by the transformer decoder have a resolution of 100$\times$100 pixels, with each pixel representing 0.5m$^2$ in the world. Our spatiotemporal model takes a 6Hz sequence of 4 images, where the final frame is the time step we make the prediction for. Our largest scale output is 100$\times$100 pixels, which we upsample to 200$\times$200 for fair evaluation with the literature. We train our network end-to-end with an Adam optimizer, batch size 8 and initial learning rate of $5e{-5}$, which we decay by 0.99 every epoch for 40 epochs.

\subsection{Ablation studies}\label{sec:ablations}
\textbf{Which way to look?} In Table \ref{tab:ablations} (top) we compare soft-attention (looking both ways), monotonic attention with lookback towards the bottom of the image (looking down) and monotonic attention with lookback towards the top of the image (looking up). The results indicate looking downwards from a point in the image is better %for generating context 
than looking upwards. This is consistent with how humans try to determine the distance of an object in an urban environment --- along with local textural clues of scale, we make use of where the object intersects the ground plane. The results also show that looking in both directions further increases accuracy, making it more discriminative for depth reasoning. 
%This indicates that looking both ways does not increase susceptibility to overfitting and is more discriminative for depth reasoning than looking only one way.%\looseness=-1

%\begin{table}%[ht]
%\centering
%\caption{IoU(\%) for baseline model vs. baseline with long-range horizontal context.}
%\label{tab:h_context}
%\begin{tabular}{cccc}
%\hline
%Model & Static classes  & Dynamic classes & Mean \\ \hline
%Baseline             & \textbf{32.4}           & \textbf{19.4}   & \textbf{25.0}                      \\
%Baseline w/ h. context             & 29.4                    & 17.3 & 22.9                  \\ \hline
%\end{tabular}\vspace{-0.2cm}
%\end{table}

%\subsection{Long-range horizontal dependencies} \label{sec:h_context}
\textbf{Long-range horizontal dependencies:} As our image-to-BEV transformation is carried out as a set of 1D sequence-to-sequence translations, a natural question is what happens when the entire image is translated to BEV (similar to `lift' approaches \cite{philion2020lift, fiery2021} 
%which use the entire image as context to determine the depth of each of its pixels
). Given the quadratic computation time and memory required to produce attention maps, 
%in addition to the quadratic memory needed to construct a matrix to store them, 
this is prohibitively expensive. However, we can approximate the contextual benefits of using the entire image by applying horizontal axial-attention on the image-plane features before the transformation. With axial-attention across the rows of the image, the pixels in the vertical scanline now have long-range horizontal context, after which we provide long-range vertical context as before by translating between 1D sequences. 

Table \ref{tab:ablations} (middle) shows that incorporating long-range horizontal context does not benefit the model and its impact is slightly detrimental. This suggests two things. Firstly, every transformed ray does not need information from the entire width of the input image, or rather, the long-range context does not provide any additional benefit over the context that has already been aggregated through the convolutions of our frontend. This indicates that performing the translation using the entire image would not increase model accuracy over the constrained formulation of our baseline. Finally, the decrease in performance from the introduction of horizontal axial-attention is possibly a sign of the difficulty in training using attention for sequences which are the width of the image; we should expect that using the entire image as the input sequence would be much harder to train.

%\subsection{Polar-agnostic vs polar-adaptive transformers}
\label{sec:polar_encs}
\textbf{Polar-agnostic vs polar-adaptive transformers:} Table \ref{tab:ablations} (bottom) compares a polar-agnostic (Po-Ag) transformer to its polar-adaptive (Po-Ad) variants. A Po-Ag model has no polar-positional information, Po-Ad in the image-plane involves polar encodings added to the transformer encoder while for the BEV-plane this information is added to the decoder. 
%As shown in Table \ref{tab:polar_variants}, 
Adding polar encodings to any one plane provides similar benefit over an agnostic model, with dynamic classes increasing the most. Adding it to both planes increases this further, but has the largest impact on static classes.

%\begin{table}%[ht]
%\centering
%\caption{IoU(\%) 
%for polar-adaptive attention mechanisms.}
%\label{tab:polar_variants}
%\begin{tabular}{cccc}
%\hline
%Model & Static classes  & Dynamic classes  & Mean \\ \hline
%Po-Ag             & 30.3	& 18.1	& 23.7                      \\
%Po-Ad (image-plane)             & 30.9 &	19.1 &	24.2                  \\
%Po-Ad (BEV-plane)            & 31.3          & 19.2   & 24.3            \\
%Po-Ad (both planes)            & \textbf{32.4}          & %\textbf{19.4}   & \textbf{25.0}            \\ \hline
%\end{tabular}
%\vspace{-0.3cm}
%\end{table}

\begin{table}
\centering
\caption{IoU(\%) for ablation studies.}
\vspace{-0.2cm}
\label{tab:ablations}
\begin{tabular}{cccc}
\hline
Model & Static classes  & Dynamic classes  & Mean \\ \hline
Looking down             & 29.5	& 15.8	& 22.1                      \\
Looking up             & 29.9 &	17.1 &	23.0                  \\
Looking both ways               & \textbf{32.4}           & \textbf{19.4}   & \textbf{25.0}            \\ \hline
Baseline             & \textbf{32.4}           & \textbf{19.4}   & \textbf{25.0}                      \\
Baseline w/ h. context             & 29.4                    & 17.3 & 22.9                  \\ \hline
Po-Ag             & 30.3	& 18.1	& 23.7                      \\
Po-Ad (image-plane)             & 30.9 &	19.1 &	24.2                  \\
Po-Ad (BEV-plane)            & 31.3          & 19.2   & 24.3            \\
Po-Ad (both planes)            & \textbf{32.4}          & \textbf{19.4}   & \textbf{25.0}            \\ \hline
\end{tabular}
\vspace{-0.5cm}
\end{table}
\color{black}

\subsection{Comparison to state-of-the-art}

\label{sec:sota_comparison}
\textbf{Baselines:}
We compare against a number of prior state-of-the-art methods. %categorised by their image-to-BEV transformation approach as either `compression' or `lift', as described in Sec.\ref{sec:related_work}. 
We begin our comparison against `compression' approaches \cite{Roddick_2020_CVPR, saha2021enabling} on nuScenes and Argoverse using the train/val splits of \cite{Roddick_2020_CVPR}. We then compare against the `lift' approach of \cite{philion2020lift, fiery2021} on nuScenes and Lyft.

In Table \ref{tab:soa_comparison}, our spatial model outperforms the current state-of-the-art compression approach of STA-S \cite{saha2021enabling} with a mean relative improvement 15\%. It is the smaller dynamic classes in particular on which we show significant improvement, with buses, trucks, trailers and barriers all increasing by a relative 35-45\%. This is supported by our qualitative results in Fig.~\ref{fig:soa_comp}, where our models show greater structural similarity to the ground truths and a better sense of shape. This difference can be partly attributed to the fully-connected layer (FCL) used in compression: when detecting small, distant objects, a large portion of the image is redundant context. Expecting the weights of the FCL to ignore redundancies to maintain only the small objects in the bottleneck is a challenge. Furthermore, 
%small distant 
objects such as pedestrians are often partially occluded by vehicles. In such cases, the FCL would be inclined to ignore the pedestrian and instead maintain the vehicle's semantics. Here the attention method shows its advantages as each radial depth can independently attend to the image
%different parts of the image 
--- so the further depths can look at the pedestrian's visible body, while depths before can attend to the vehicle. Our results on the Argoverse dataset in Table \ref{tab:argo_results} demonstrate similar patterns, where we improve upon PON \cite{Roddick_2020_CVPR} by a relative 30\%. 

In Table \ref{tab:us_vs_philion} we outperform LSS \cite{philion2020lift} and FIERY \cite{fiery2021} on nuScenes and Lyft (FIERY \cite{fiery2021} uses the `lift' approach of \cite{philion2020lift}). A true comparison on Lyft is not possible as it doesn't have a canonical train/val split and we were unable to acquire those used by \cite{philion2020lift}. While we used splits of similar sizes to \cite{philion2020lift}, the exact scenes are unknown. As a `lift' approach bears some similarity to our translation approach in that the network is able to select how to distribute image context across its polar ray, the difference in performance here can likely be attributed to our constrained, spatially-aware translations between scanlines and rays. One of the avenues for future work is improving localisation accuracy for distant objects, and their false negatives.
Finally, our approach is easily transferrable to indoor mobile robotics applications once ground truth has been collected to train the models.

\begin{table}%[ht]
\centering
\caption{IoU(\%) on the Argoverse validation split of \cite{Roddick_2020_CVPR}.}
\vspace{-0.2cm}
\label{tab:argo_results}
\resizebox{\columnwidth}{!}{%
\begin{tabular}{l|cccccccc|c}
     & Driv.         & Veh.          & Ped.         & L.Veh.        & Bic.         & Bus.          & Trail.        & Mot. & Mean          \\ \hline
PON \cite{Roddick_2020_CVPR}  & 65.4          & 31.4          & \textbf{7.4} & 11.1          & 3.6 & 11            & 0.7           & 5.7  & 17.0          \\
Ours & \textbf{75.9} & \textbf{35.8} & 5.7          & \textbf{14.9} & \textbf{3.7}          & \textbf{30.2} & \textbf{12.2} & 2.6  & \textbf{22.6} \\ \hline
\end{tabular}%
}
\vspace{-0.3cm}
\end{table}

\begin{table}%[ht]
\centering
\caption{IoU(\%) for spatial (S)/spatiotemporal (ST) methods.}
\vspace{-0.2cm}
\label{tab:us_vs_philion}
\begin{tabular}{l|ccc|ccc}
           & \multicolumn{3}{c|}{nuScenes} & \multicolumn{3}{c|}{Lyft}                \\ \cline{2-7} 
           & Driv.     & Car     & Veh.    & Driv. & Car  & \multicolumn{1}{l|}{Veh.} \\ \hline
(S) LSS    & 72.9      & 32.0    & 32.0    & -     & 43.1 & \multicolumn{1}{l|}{44.6} \\
(S) FIERY  & -         & 37.7    & -       & -     & -    & \multicolumn{1}{l|}{-}    \\
(S) Ours   & \textbf{78.9}      & \textbf{39.9}    & \textbf{38.9}    & \textbf{82.0}  & \textbf{45.9} & \multicolumn{1}{l|}{\textbf{45.4}} \\ \hline
(ST) FIERY & -         & 39.9    & -       &       &      &                           \\
(ST) Ours  & \textbf{80.5}      & \textbf{41.3 }   & \textbf{40.2}    &       &      &                           \\ \cline{1-4}
\end{tabular}
\vspace{-0.5cm}
\end{table}

% \begin{table}%[ht]
% \vspace{-0.3cm}
% \centering
% \caption{IoU(\%) against spatiotemporal methods on nuScenes.}
% \vspace{-0.3cm}
% \label{tab:us_vs_fiery}
% \begin{tabular}{l|ccc|}
%         & \multicolumn{3}{c|}{nuScenes} \\
%         & Driv.     & Car     & Veh.    \\ \hline
% FIERY\cite{fiery2021}   & -         & 39.9    & -       \\
% Our S.T & \textbf{80.5}      & \textbf{41.3}    & \textbf{40.2}   
% \end{tabular}
% \vspace{-0.3cm}
% \end{table}

\vspace{-0.1cm}
\section{Conclusion} \label{sec:conclusion}
We proposed a novel use of transformer networks to map from images and video sequences to an overhead map or bird's-eye-view of the world. 
We combine our physical-grounded and constrained formulation, with ablation studies that make use of progress in monotonic attention to confirm our intuitions whether context above or below a point is more important for this form of map generation.
Our novel formulation obtains state-of-the-art results for instantaneous mapping of three well-established datasets.
\vspace{-0.2cm}
\section*{Acknowledgements}
This project was supported by the EPSRC project ROSSINI (EP/S016317/1) and studentship 2327211 (EP/T517616/1).

\bibliographystyle{IEEEtran}
\bibliography{IEEEabrv,ref}

\end{document}